\documentclass{article}

\usepackage{arxiv}

\usepackage[utf8]{inputenc} 
\usepackage[T1]{fontenc}    
\usepackage{hyperref}       
\usepackage{url}            
\usepackage{booktabs}       
\usepackage{amsfonts}       
\usepackage{nicefrac}       
\usepackage{microtype}      
\usepackage{lipsum}
\usepackage{graphicx}
\usepackage{subfigure}
\usepackage{multirow}

\usepackage{epigraph}

\graphicspath{ {./images/} }

\title{ERNIE-ViLG: Unified Generative Pre-training for Bidirectional Vision-Language Generation}

\author{\textbf{Han Zhang} \and \textbf{Weichong Yin} \and \textbf{Yewei Fang} \and \textbf{Lanxin Li} \\
\and \textbf{Boqiang Duan} \\ 
\and \textbf{Zhihua Wu~~~~~~~~~~~~~~~Yu Sun~~~~~~~~~~~~~~~~Hao Tian~~~~~~~~~~~~~~~~Hua Wu~~~~~~~~~~~~~~~Haifeng Wang}\\ \\
\bf{ Baidu Inc}. \\ \\
 \{\texttt{zhanghan17, yinweichong, sunyu02}\}\texttt{@baidu.com} \\
}

\begin{document}
\maketitle
\begin{abstract}
Conventional methods for the image-text generation tasks mainly tackle the naturally bidirectional generation tasks separately, focusing on designing task-specific frameworks to improve the quality and fidelity of the generated samples. 
Recently, Vision-Language Pre-training models have greatly improved the performance of the image-to-text generation tasks, but large-scale pre-training models for text-to-image synthesis task are still under-developed.
In this paper, we propose ERNIE-ViLG, a unified generative pre-training framework for bidirectional image-text generation with transformer model. Based on the image quantization models, we formulate both image generation and text generation as autoregressive generative tasks conditioned on the text/image input. 
The bidirectional image-text generative modeling eases the semantic alignments across vision and language. 
For the text-to-image generation process, we further propose an end-to-end training method to jointly learn the visual sequence generator and the image reconstructor. 
To explore the landscape of large-scale pre-training for bidirectional text-image generation, we train a 10-billion parameter ERNIE-ViLG model on a large-scale dataset of 145 million (Chinese) image-text pairs which achieves state-of-the-art performance for both text-to-image and image-to-text tasks, obtaining an FID of 7.9 on MS-COCO for text-to-image synthesis and best results on COCO-CN and AIC-ICC for image captioning. 
\end{abstract}

\section{Introduction}
\epigraph{``What I cannot create, I do not understand.''}{\textit{Richard Feynman}}
Cross-modal generation, aiming at mapping or translating one modality to another, requires the model to fully "understand" the source modality and to faithfully "create" (generate) the target modality with high semantic consistency with the source  \cite{baltruvsaitis2018multimodal}. As concerned with a big part of multimodal machine learning researches, the cross-modal generation task has attracted dramatic attentions, for audio-linguistic \cite{hinton2012deep,oord2016wavenet}, visual-linguistic \cite{vinyals2015show,reed2016generative,anderson2018bottom,zhou2020unified,zhang2021cross,ramesh2021zero},  audio-visual \cite{chen2017deep,zelaszczyk2021audio} modalities. In this paper, we focus on the visual-linguistic generation tasks which mainly include image captioning (describing the visual content of an image in natural language) and text-to-image synthesis (creating an image that keeps semantic consistent with a textual description). As the intersection of computer vision and natural language processing, the Vision-Language generation tasks attract much interest from both NLP and CV communities and have achieved great progresses over the past several years. 

Naturally, the text-image generation tasks are bidirectional and a reasonable model should have the capability of both image captioning and text-to-image synthesis. 
However, due to the distinct generative architectures for language generation and image generation, methods tackle the two tasks have developed separately. 
The conventional methods for image captioning mainly adopt 
the encoder-decoder architecture \cite{vinyals2015show} where the text is generated in a sequence generation manner. The image is first encoded into one or multiple vectors and then a decoder (e.g., LSTM) is utilized to generate language tokens based on the encoder results.  Recently, the transformer-based Vision-Language Pre-trained models \cite{zhou2020unified,li2020oscar} have significantly improved the performance of image captioning benefiting from pre-training on large-scale image-text aligned datasets.
On the other hand, the dominant text-to-image synthesis methods are the Generative Adversarial Networks (GANs) \cite{goodfellow2014generative} variants where ConvNets are utilized as the basic architecture of the generator. While various methods have been proposed to improve the quality and resolution of generated images, such as StackGAN \cite{zhang2017stackgan}, XMC-GAN \cite{zhang2021cross}, it remains difficult to generate images of complex scenes with multiple objects. 

Inspired by the autoregressive models for the images \cite{oord2016conditional,chen2020generative}, recent works \cite{ramesh2021zero,ding2021cogview} have proposed to formulate the text-to-image synthesis as a sequence-to-sequence problem where the image tokens are learned through discrete VAE \cite{oord2017neural,ramesh2021zero}. They model the text tokens and image tokens in the same transformer and have shown great improvement for the image quality and the capability to construct complex scenes. 
And the text-to-image generation process usually adopts a two-stage pipeline consisting of a generator (to generate the image tokens) and a reconstuctor (to build the image from the generated image tokens), which are trained separately.

More recently, researchers have attempted to unify the bidirectional image-text generation tasks in a single model. Cho {\em et\ al.} \cite{cho2020x} attempt to enhance the existing Vision-Language Pre-training models with the capability of text-to-image generation, utilizing image discrete tokens generated by K-means clustering of pre-trained image features and generating the image sequence in a non-autoregressive manner. 
Based on this approach, Huang {\em et\ al.} \cite{huang2021unifying} formulate both the tasks as sequence generation tasks and further propose a sequence-level training task to improve the performance. 
Although they unify the two tasks in one single transformer, they employ the non-autoregressive sampling to generate images which enables a fast inference speed but leads to suboptimal performance compared to the autoregressive generative models due to the removal of explicitly modeling of dependencies among the target tokens. 

In the paper, we propose ERNIE-ViLG, a unified pre-training method for the bidirectional image-text generation with transformer model. Based on image quantization techniques \cite{oord2017neural,ramesh2021zero,esser2021taming}, both the image-to-text and text-to-image generation are tackled in autoregressive manners within the same transformer.
And we further propose an end-to-end pre-training method for text-to-image synthesis which makes the traditionally separate two-stage generator and reconstructor training jointly. 
To explore the landscape of large-scale pre-training for bidirectional text-image generation, we pre-train a 10-billion parameter model on a large-scale dataset of 145 million high-quality Chinese image-text pairs. It has shown superior performance for both text-to-image synthesis and image captioning tasks. Concretely, it achieves the best FID on MS-COCO for text-to-image synthesis and obtains state-of-the-art performance on two Chinese image captioning datasets. Moreover, we evaluate ERNIE-ViLG on a challenging generative Visual Question Answering (VQA) task, and the excellent performance shows that our bidirectional generative model has captured the semantic alignments across vision-language modalities and can also be transferred to tackle complex generative tasks besides image-text generation.

Overall, our contributions include:
\begin{itemize}
\item We propose ERNIE-ViLG, a unified generative pre-training method for bidirectional image-text generation tasks, where both the image and text generation are formulated as autoregressive generative tasks. And we propose the first end-to-end training method for text-to-image synthesis based on image discrete representation, which enhances both the generator and reconstructor and outperforms the traditional two-stage approach.
\item We train a 10-billion parameter ERNIE-ViLG model and obtain superior performance for both text-to-image and image-to-text generation tasks, setting new SOTA results for text-to-image synthesis on MS-COCO and obtaining SOTA results for image captioning on two popular Chinese datasets.
\item Superior performance on the generative VQA task shows that our bidirectional generative model  captures the complex semantic alignments between the vision and the language modalities. 
\end{itemize}

\section{Related Work}
\subsection{Vision-Language Pre-training}
Vision-Language Pre-training (VLP) models \cite{lu2019vilbert,zhou2020unified,chen2019uniter,yu2020ernie,gan2020large,li2020unimo,zhang2021vinvl} have greatly improved the performance of various Vision-Language tasks, such as VQA and cross-modal retrieval. 
Utilizing transformer architecture (two-stream or single-stream) to fuse the visual modality and the linguistic modality, they propose various pre-training tasks, from the conventional masked language modeling (MLM), masked region prediction (MRP), image-text matching (ITM) to word-region alignment \cite{chen2019uniter}, scene graph prediction \cite{yu2020ernie}, multimodal conditional text generation \cite{cho2021unifying}, Prefix-LM \cite{wang2021simvlm}, etc. The pre-training datasets are also a big concern of VLP researches where early works utilize million-scale noisy image-text datasets such as Conceptual Captions \cite{sharma2018conceptual} and human labeled datasets such as COCO Captions \cite{chen2015microsoft} and recent works scale up the pre-training with hundreds of millions and nearly billion-scale noisy image-text dataset \cite{wang2021simvlm}.

While most of VLP methods focus on pre-training for the vision-language understanding tasks, some works have noticed the importance of pre-training for improving the performance of cross-modal generation tasks, mainly image captioning task. Zhou {\em et\  al.} \cite{zhou2020unified} propose the first unified model to improve the performance of both understanding and generation tasks, where the bidirectional and sequence-to-sequence pre-training tasks utilize different attention masks. Wang {\em et\ al.} \cite{wang2021simvlm} propose prefix-LM pre-training task, similar to image captioning, solely exploiting language modeling objective. Hu {\em et\ al.} \cite{hu2021scaling} study the scaling behavior of pre-training from both the data and model perspective for image captioning.

While the image-to-text generation task has been advanced a lot due to the pre-training on large-scale image-text datasets, the benefit of pre-training for text-to-image synthesis task has not been fully explored. Cho {\em et\ al.} \cite{cho2020x} propose to enhance the VLP models  with image generation capability based on discretized visual representation and fine-tuned for text-to-image synthesis. Wu {\em et\ al.} \cite{wu2021n} attempt to construct a unified encoder-decoder pre-trained framework aiming at both image and video synthesis tasks. 

ERNIE-ViLG focuses on the pre-training for both the image-to-text and text-to-image generation tasks and simplifying the pre-training objectives to the autoregressive sequence-to-sequence generation tasks for both image and text.

\subsection{Vision-Language Generation}
\paragraph{Image-to-Text Generation}
The typical methods for image captioning adopt the encoder-decoder architecture formulating the image-to-text problem as text sequence generation task with the context of the image. Vinyals {\em et\ al.}  \cite{vinyals2015show} propose a simple LSTM-based  architecture where the image is encoded into a single vector and used as the initial hidden state of a single-layer LSTM. Utilizing the attention-mechanism, various methods \cite{xu2015show,anderson2018bottom,huang2019attention} have been proposed to improve the correlation between the visual image and the generated text sequence where the image is encoded into multiple features and used via cross-modal attention to guide the generation process. 
Recently, researchers have explored the potential of transformer-based model for image captioning \cite{sharma2018conceptual,herdade2019image,zhou2020unified}. 
Zhou {\em et\ al.} \cite{zhou2020unified} propose to use a unified transformer architecture to fuse the visual and textual modalities, for both encoding and decoding. Following this, Li {\em et\ al.} \cite{li2020oscar} propose to include objects tags which are extracted from the image using an object detector into the input for augmenting the semantic alignment between the image and text.

\paragraph{Text-to-Image Synthesis}
Since the arise of generating images using Generative Adversarial Networks (GANs) \cite{goodfellow2014generative}, the GANs based methods have been most popular ones for text-to-image synthesis. Reed {\em et\ al.} \cite{reed2016generative} first extend the conditional GANs to generate images from language descriptions. After that, Zhang {\em et\ al.}  \cite{zhang2017stackgan} propose StackGAN to generate high-resolution images in a multi-stage manner. Zhu {\em et\ al.} \cite{zhu2019dm} propose to use dynamic memory networks to refine image contents. Zhang {\em et\ al.} \cite{zhang2021cross} propose to incorporate contrastive learning to improve the fidelity of the generated images to the textual input. Recently, inspired by the autoregressive generative models for pixel-by-pixel image generation \cite{chen2020generative}, the transformer-based methods \cite{ramesh2021zero,ding2021cogview} have shown promising results for text-to-image synthesis. Based on the discrete image tokens learned by various discrete VAE \cite{oord2017neural,esser2021taming}, they try to model the image tokens and text tokens in a single transformer framework, following a unidirectional manner for both the text input and the image target. 

\section{Approach}
In this section, we will introduce our unified generative pre-training framework for bidirectional Vision-Language generation and the end-to-end method for text-to-image synthesis training. Also, we present the details of the collection of the large-scale image-text dataset and the distributed training strategies for pre-training the 10-billion parameter model.
\subsection{A Unified Generative Framework for Bidirectional Image-text Generation}
\begin{figure}[htb]
  \centering
  \includegraphics[width=0.9\textwidth]{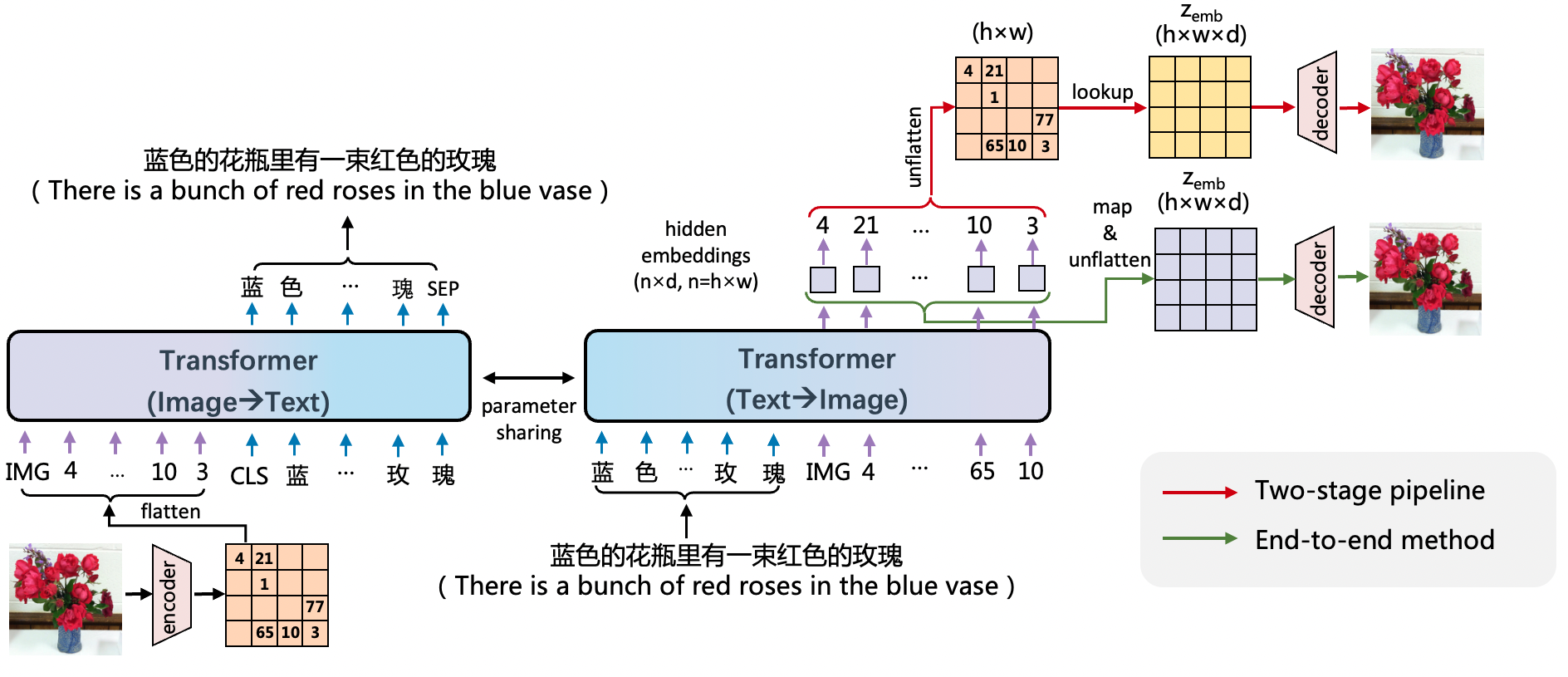}
  \caption{The unified model architecture of ERNIE-ViLG for bidirectional image-text generation.}
  \label{fig:bimodel}
\end{figure}

As shown in Figure \ref{fig:bimodel}, ERNIE-ViLG adopts a unified framework for the bidirectional image-text generation. The image is represented as a sequence of discrete representation by vector quantization variational autoencoder (VQVAE) \cite{oord2017neural}. 
The image discrete sequence is used as the input(output) of a parameter-sharing transformer for autoregressive image-to-text (text-to-image) generation.
Specifically, for the image-to-text generation, the transformer takes the image discrete sequence as input to generate the corresponding textual sequence. Conversely, for the text-to-image synthesis, the text is inputted to the transformer for generating the corresponding visual discrete sequence, and then the image discrete sequence is used for reconstructing the image. Besides the traditional two-stage pipeline approach, we further propose an end-to-end training method of jointly training the modules for discrete representation sequence generation and image reconstruction to strengthen the ability of text-to-image synthesis.
\subsubsection{Image Discrete Representation}
Discrete representation has achieved significant improvements for image generation tasks recently, because it can be better adapted to autoregressive modeling. VQVAE \cite{oord2017neural} proposes to represent images as latent discrete sequences using vector quantization. It has stronger semantic representation ability than pixel clustering. VQVAE encodes the original image $x \in \mathbb{R}^{H \times W \times 3}$ by encoder \textit{E} and quantizes it to $z$ with codebook \textit{C}. $z$ is the discrete sequence with $h \times w$ indices, and used to reconstruct $x$ via decoder $G$ and $C$. The loss function during training is shown in Equation 1-2.
\begin{equation}
\mathcal{L} _{VQVAE}=\parallel x-\hat{x} \parallel^2+\parallel sg[z_{emb}] - E(x)\parallel^2_2 +\parallel sg[E(x)] - z_{emb}\parallel^2_2
\end{equation}
\begin{equation}
\hat{x}=G(z_{emb})= G(C(z)) = G(q(E(x)))
\end{equation}

The overall loss function contains two parts. The first term of $\mathcal{L}_{VQVAE}$ is reconstruction loss. The last two terms aim to make \textit{E(x)} (the encoder outputs) and $z_{emb}$ (the embeddings after the lookup of discrete sequence from codebook) close to each other.  \textit{sg[·]} denotes the stop-gradient operation, and \textit{q[·]} denotes the vector quantization. \textit{x} and $\hat{x}$ are the original and reconstructed images. The discrete sequence \textit{z} servers as the input (or output) of the bidirectional generative model, and its length \textit{n} is $h \times w$.

\subsubsection{Bidirectional Generative Model}
For the generative model, we use a multi-layer transformer encoder for both image-to-text and text-to-image generation tasks. 
Different from the decode-only \cite{ramesh2021zero, ding2021cogview} or separate encoder-decoder \cite{wu2021n} architectures, the encoder and decoder of ERNIE-ViLG share parameters and use specific self-attention masks to control the contexts like UniLM \cite{dong2019unilm} and ERNIE-GEN \cite{xiao2020ernie-gen}, where the source tokens are allowed to attend all the source tokens and the target tokens are allowed to attend the source tokens and the target tokens lie left to them. The bidirectional generation pre-training tasks are modeled exact in the same model where we consider sharing of model space helps establish better semantic alignments across vision and language modalities.

The visual tokens $\left\{z_1,…,z_n\right\}$ discretized from the image using VQVAE encoder and the textual tokens $\left\{t_1,…,t_m\right\}$ tokenized from the text using WordPiece tokenizer are concatenated and fed into the transformer model. Therefore, during training, the model takes the input stream of $[t_1,…,t_m,z_1,…,z_n]$ for text-to-image generation task to predict the image tokens autoregressively and $[z_1,…,z_n,t_1,…,t_m]$ for image-to-text generation task. 
We use the following multi-task loss for learning the image-text bidirectional generation tasks.
\begin{equation}
\mathcal{L}=\mathcal{L} _{txt2img} + \mathcal{L} _{img2txt} 
\end{equation}
\begin{equation}
\mathcal{L} _{txt2img} = \sum_{k=1}^{n}{-\log{P(z_k|t_1,…,t_m,z_1,…,z_{k-1})}}
\end{equation}
\begin{equation}
\mathcal{L} _{img2txt} = \sum_{k=1}^{m}{-\log{P(t_k|z_1,…z_n,t_1,…,t_{k-1})}}
\end{equation}

\begin{figure}[htb]
    \centering
    \subfigure[Sparse attention for text-to-image generation]{
        \includegraphics[width=0.9\linewidth]{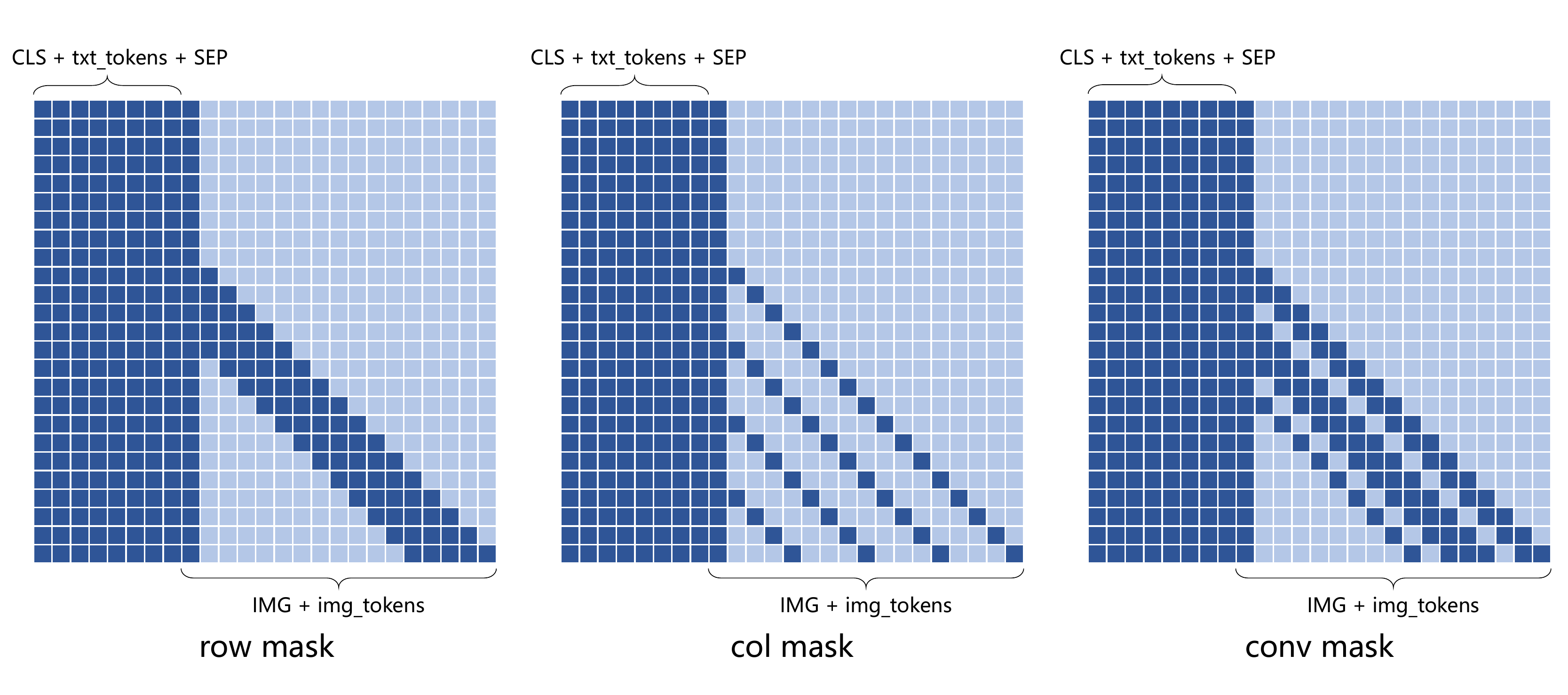}
    }
    \subfigure[Sparse attention for image-to-text generation]{
        \includegraphics[width=0.9\linewidth]{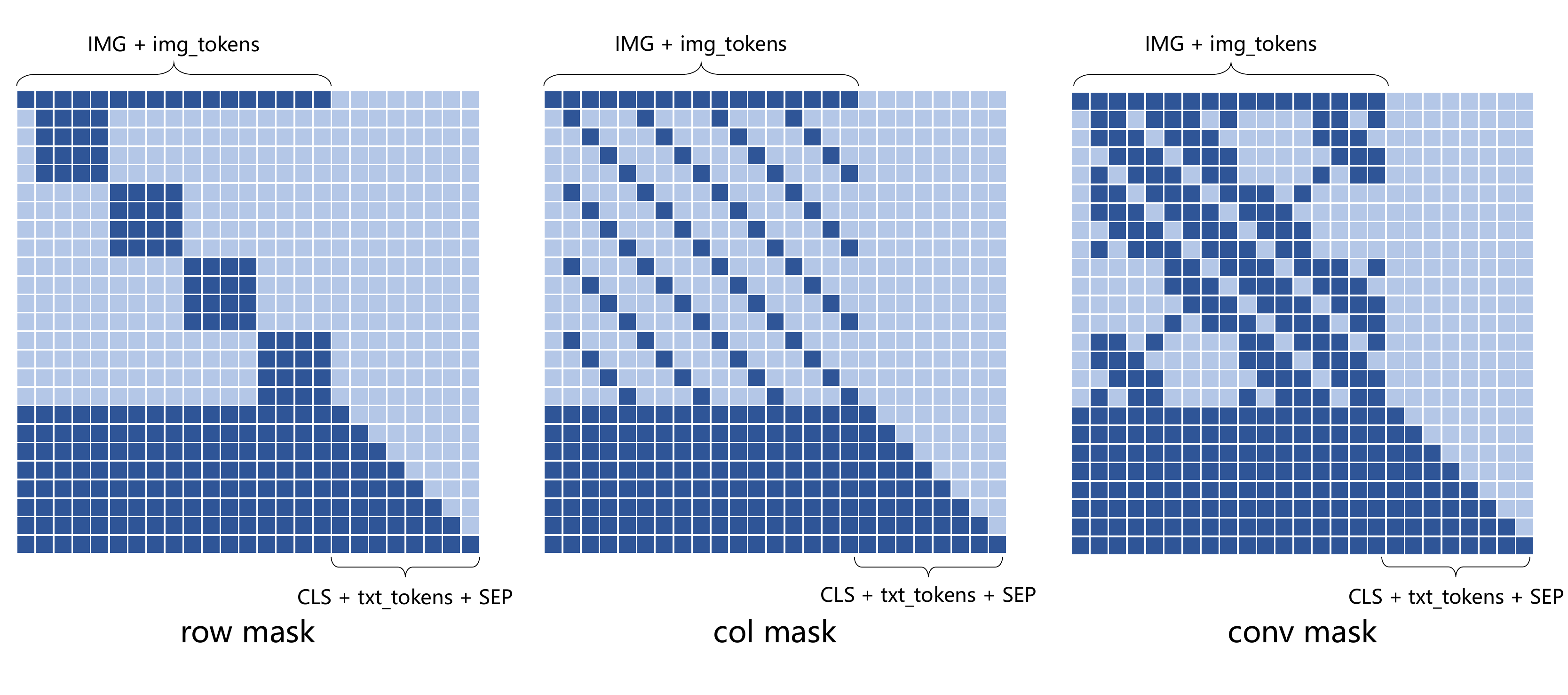}
    }
    \caption{Sparse attention for bidirectional text-image generation. For text-to-image generation, text tokens are visible to each other and the image attention pattern is as same as DALL-E. But for image-to-text generation, the attention pattern of image tokens in the same row, column or convolutional kernel is bidirectional rather than unidirectional. 
    This is a hypothetical version of our transformer with a maximum text length of 6 tokens, image length of 16 tokens (4×4), and convolutional kernel size of 3×3. The actual kernel size of our model is 11×11 following DALL-E.}
    \label{fig:sparse_attention}
\end{figure}

Note that the length $n$ of visual sequence $z$ is often large (usually larger than 1024) to reduce the information loss of the image, which causes a relatively high computation cost and memory consumption for the transformer model during training and inference. As shown in Figure \ref{fig:sparse_attention}, we utilize a similar sparse attention mechanism as in \cite{ramesh2021zero}. Concretely, we adopt row attention (i mod 4 != 2 ),  column attention (i mod 4 = 2) and convolutional attention (the last layer) for the i-th transformer layer. We implement the row attention and column attention as block-wise attention while training. Several primary experiments verify that this sparse attention implementation can roughly increase the speed by 25\% and save 50\% of GPU memory during training while keeping the convergence of loss consistent with dense attention. And the improvement for predicting is much more significant. While the sparse attentions in \cite{ramesh2021zero} are designed for unidirectional modeling, we adapt them to the bidirectional modeling manner of the visual input tokens for the image-to-text generation as shown in Figure \ref{fig:sparse_attention}(b). 

\subsubsection{Text-to-Image Synthesis}
Text-to-image synthesis based on image discrete sequence is usually tackled in a two-stage pipeline: discrete representation sequence generation and image reconstruction. As the red path in Figure \ref{fig:bimodel}, the generated discrete sequence is looked up in the codebook to obtain a 3D tensor $z_{emb} \in \mathbb{R}^{h \times w \times d}$. Then $z_{emb}$ is sent to the reconstructed decoder to be restored into an image. The generator and the reconstructor are trained independently. 

Besides the traditional two-stage pipeline mode, in the framework of ERNIE-ViLG, the text-to-image synthesis can also use our newly proposed end-to-end training method as the green path in Figure \ref{fig:bimodel}. Hidden embeddings of image tokens outputted by the last transformer layer are mapped to $z_{emb}$ nonlinearly. Gradients could be propagated backward from the reconstructor to the generator since the non-derivable ID mapping operation is avoided. Therefore, the end-to-end training can be carried out.

This method is designed to:
\begin{itemize}
\item provide more contextual features for the reconstructor. Compared with context-independent embedding in codebooks, hidden embedding is encoded by a deep model and contains more image semantics. It also has perception of the textual information through the attention interaction.
\item enhance the generator with the reconstruction task. The hidden embedding receives both abstract and original supervised signals from generation and reconstruction. It helps the generator learn better about image representation.
\end{itemize}

Our experiments show that the end-to-end training can improve both generator and reconstructor compared with the two-stage pipeline. Detailed experimental results are presented in Section \ref{section:end2endexp}.

\subsubsection{Image-to-Text Generation}
For the image-to-text generation task, the image is first discretized into visual tokens using the encoder and the codebook of the VQVAE. The image tokens are then fed into the Transformer to generate the text tokens autoregressively. In our current implementation, the quantization modules are pre-trained and fixed during the image-to-text generation training task while they could also be updated jointly with the generative model, and we will explore it for the future work.

\subsection{Large-scale Pre-training of ERNIE-ViLG}
To explore the landscape of large-scale pre-training for bidirectional text-image generation, we train a 10-billion parameter ERNIE-ViLG model, a 48-layer transformer encoder for both image-to-text and text-to-image generation. The core issues of large-scale pre-training include collection of training data and distributed parallel training.

\subsubsection{Large-scale Image-Text Dataset Collection}
To pre-train a generative model with general capabilities of bidirectional text-image generation, from basic entities to complex scenes, we build a large-scale image-text dataset consisting of over 145 million high-quality Chinese image-text pairs. The sources of our dataset are listed as follows: 
\begin{itemize}
\item \textbf{Chinese Webpages}. We crawl 800 million raw Chinese alt-text descriptions paired with images from various Chinese webpages, conduct several steps of filtering and totally harvest 70 million text-image pairs. The filtering rules mainly include: (1) Text-length: the number of words in alt-text is less than 15. (2) Text-content: the alt-text must contain at least one noun and contain no special characters. (3) Image-text similarity: the similarity score of between the alt-text and image (calculated by an in-house text-image matching model with the score range from 0.0 to 1.0) is greater than 0.5.
\item \textbf{Image Search Engine}. We collect roughly 60 million query texts and corresponding user-clicked images from our internal image search engine. There is often a strong correlation between the query and user-clicked images.
\item \textbf{Public image-text Dataset}. We collect a total of 15 million text-image pairs from two public datasets, CC \cite{sharma2018conceptual} and CC12M \cite{changpinyo2021conceptual}. The captions in these datasets are translated to Chinese through Baidu Translate API. \end{itemize}

\begin{figure}[htb]
  \centering
  \includegraphics[width=0.9\textwidth]{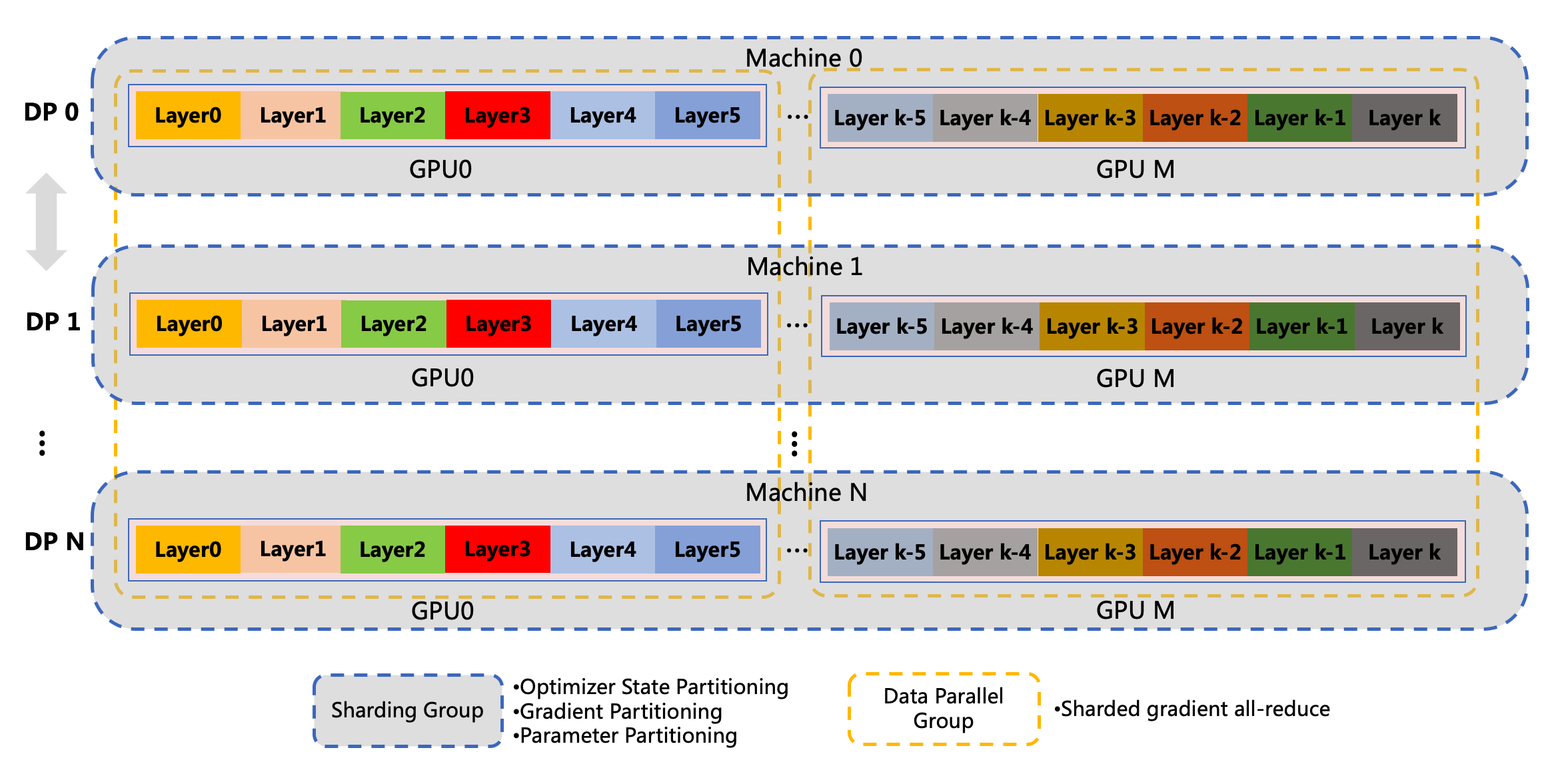}
  \caption{Hybrid parallelism of data parallelism and sharding in PaddlePaddle. We organize the cluster in two dimensions: the data is sharded in column dimension while the model states (optimizer states, gradients, and parameters) are sharded among row dimensions. All communication needed by sharding is confined to the same machine.}
  \label{fig:paddledist}
\end{figure}

\subsubsection{Distributed Training Strategies for Large-Scale Generative Models}
The 10-billion parameter ERNIE-ViLG is implemented based on PaddlePaddle platform \cite{ma2019paddle}. Serious challenges need to be addressed to train such a large-scale model, such as limited device memory and computation efficiency. Also, it is rather demanding to fit a 10B generative model into a single GPU, and it becomes more challenging considering our bidirectional model’s structure which doubles the memory consumption of activations and gradients. 

As shown in Figure \ref{fig:paddledist}, we adopt Group Sharded data parallelism techniques \cite{rajbhandari2020ZeRO, yulong2021adaptive} to eliminate memory redundancies by partitioning the optimizer states, gradients, and parameters across multiple devices. In addition, activation recomputation \cite{chen2016recomputation} and mixed-precision \cite{micikevicius2017amp} are applied to reduce the GPU memory footprint and increase throughput. Moreover, Optimizer-offload \cite{rajbhandari2021ZeRO-offload} is introduced to swap the sharded optimizer states and master parameters to the CPU which largely reduces the GPU memory footprint. Furthermore, we combine Optimizer-offload with Gradient Accumulation to bring down the communication frequency between CPU and GPU, delivering a higher computation efficiency.

\section{Experiments}
 We pre-train a 10-billion parameter ERNIE-ViLG model and verify its generation ability. We conduct experiments on the conventional bidirectional image-text tasks: text-to-image synthesis and image captioning. Besides, in order to further verify the cross-modal understanding ability of our model, we transfer ERNIE-ViLG for the challenging generative VQA task to evaluate the alignments our model has captured across vision and language modalities. 

\subsection{Settings}
We use VQGAN \cite{esser2021taming}, an enhanced variant of VQVAE, as our "image tokenizer". By adding adversarial training and perceptual loss, VQGAN enables clearer and more realistic image restoration from discrete representations. We adopt the model of VQGAN with f = 8 and vocab size = 8192, where f denotes the reduction factor in the side-length. We pre-process the original image to 256 * 256 through center crop, and thus the length of the visual discrete token sequence \textit{n} is 1024 ($h \times w, h=w=32$). 

For the generator, we use a multi-layer transformer encoder for both image-to-text and text-to-image generation tasks, which consists of 48 transformer layers with 4096 hidden units and 64 attention heads and totally has 10 billion parameters. Considering the instability of the training of GAN-based models, we choose the two-stage text-to-image synthesis mode when training this model and use the decoder of VQGAN as the image reconstructor directly.

\subsection{Text-to-Image Synthesis}
To generate images given the text, we follow the same sampling strategy as in \cite{ramesh2021zero} and the generated images are re-ranked with an in-house contrastive learning-based image-text matching model. We carry out the automatic evaluations of ERNIE-ViLG on a commonly used text-to-image synthesis dataset in both zero-shot and fine-tuning settings. Moreover, we conduct a human evaluation campaign to directly assess the quality of the images generated. 
\paragraph{Automatic Evaluation}

\begin{table}[htb]
 \caption{Comparison of FID with previous text-to-image synthesis models on MS-COCO. Results of other models are directly taken from their papers. "Pre-trained" means the model is obtained by vision-language pre-training, "Supervised" illustrates the model is trained in a fully-supervised manner.}
  \centering
  \begin{tabular}{l|c}
    \toprule
    Model & FID $\downarrow$ \\
    \midrule
    \multicolumn{2}{c}{Supervised (fine-tuned)}\\
    \toprule
    AttnGAN \cite{xu2018attngan} & 35.5 \\
    DM-GAN \cite{zhu2019dm} & 32.6 \\
    DF-GAN \cite{tao2020df} & 21.4 \\
    XMC-GAN \cite{zhang2021cross} & 9.3 \\
    X-LXMERT \cite{cho2020x} & 37.4 \\ 
    Huang {\em et\ al.} \cite{huang2021unifying} & 29.9 \\
    NÜWA \cite{wu2021n} & 12.9 \\
    ERNIE-ViLG & \textbf{7.9} \\
    \bottomrule
    \multicolumn{2}{c}{Zero-shot}\\
    \toprule
    DALL-E \cite{ramesh2021zero} & 27.5 \\
    CogView \cite{ding2021cogview} & 27.1 \\
    ERNIE-ViLG & \textbf{14.7} \\
    \bottomrule
  \end{tabular}
  \label{exp-zero-t2i-auto-result}
\end{table}

For automatic evaluation, we compare ERNIE-ViLG with other strong methods on MS-COCO \cite{lin2014microsoft}. MS-COCO is a publicly available benchmark for text-to-image synthesis, which is challenging for containing many complex scenes that involve common objects. Following previous works, we randomly sample 30,000 images from the validation set and translate their corresponding captions to Chinese through Baidu Translate API. For the generated images, we rerank the samples and select best of 60 samples for the zero-shot experiments for a fair comparison with \cite{ding2021cogview} and best of 10 samples when fine-tuning. Fr\'{e}chet Inception Distance (FID) \cite{heusel2017gans} is adopted for image quality assessment\footnote{Same as previous works, we adopt evaluation code from \href{https://github.com/MinfengZhu/DM-GAN}{https://github.com/MinfengZhu/DM-GAN}}.

The results illustrated in Table \ref{exp-zero-t2i-auto-result} compare our model with previous works. In the zero-shot setting, ERNIE-ViLG surpasses DALL-E which is a 12-billion parameter model by a large margin, with a significant FID improvement of 12.8, even comparable to the fully-supervised models. It confirms that ERNIE-ViLG acquires a general semantic alignment between image and text, even for complex open-domain scenes. After fine-tuning on the domain-specific dataset, ERNIE-ViLG achieves the state-of-art result on MS-COCO, improving the FID metric by 5.0 and 1.4 over the best transformer-based method and the best GAN-base method respectively.

\paragraph{Human Evaluation}
\begin{table}[htb]
 \caption{Average scores (1\textasciitilde5) of human evaluation.}
  \centering
  \begin{tabular}{l|c|c|c}
    \toprule
    Model & Image Clarity & Texture Quality & Relevance to the Text \\ \midrule
    CogView     & 3.867 & 2.623 & 2.203  \\  
    ERNIE-ViLG  & \textbf{4.221} & \textbf{2.723} & \textbf{2.641}  \\
    \bottomrule
  \end{tabular}
  \label{exp-zero-t2i-human-result}
\end{table}

To obtain a direct assessment of the quality of images our model generated in zero-shot setting, we build a diverse dataset for human evaluation, consisting of 500 texts for various circumstances. 
The text sentences are collected from a variety of aspects to fully explore the general capability of ERNIE-ViLG, such as detailed attributes for objects, combining multiple objects in a reasonable manner, etc. See more details about this dataset in Appendix \ref{sec: t2i-human-eval-dataset}. 
For the evaluations, three evaluators are asked to assess each image from three perspectives (image clarity, texture of the image, relevance between the text and the image) with a quality score range from 1 to 5. We rerank and select the best of 60 generated samples for each text and compare against CogView\footnote{Given texts, images generated by CogView are manually crawled from its official website.}. 
The average evaluation results are listed in Table \ref{exp-zero-t2i-human-result}. ERNIE-ViLG achieves higher quality scores than CogView \cite{ding2021cogview} from all three perspectives which shows that our model acquires better zero-shot text-to-image generation capability.

\begin{figure}[htb]
    \centering
    \subfigure[]{
        \includegraphics[width=0.15\linewidth]{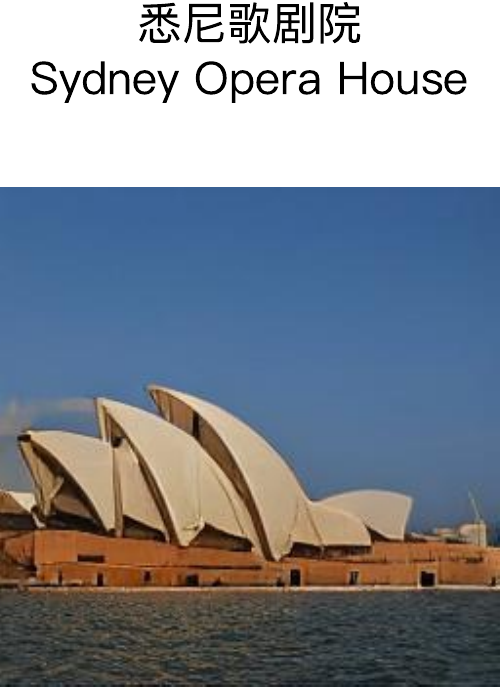}
        \label{fig:t2i_case_openDomain_sim1}
    }
    \subfigure[]{
        \includegraphics[width=0.15\linewidth]{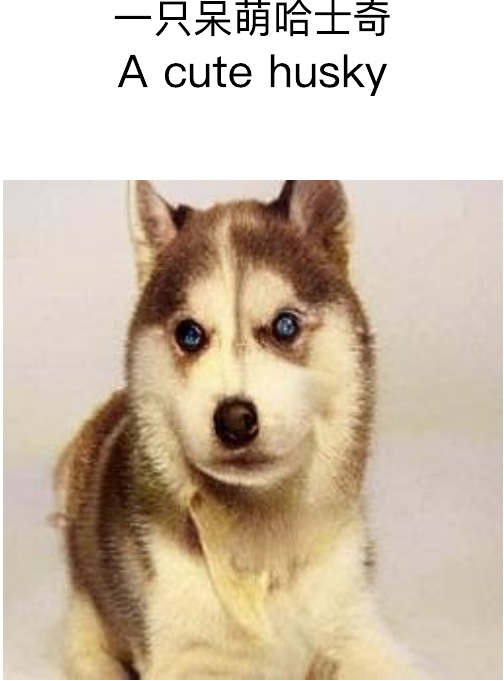}
        \label{fig:t2i_case_openDomain_sim2}
    }
    \subfigure[]{
        \includegraphics[width=0.15\linewidth]{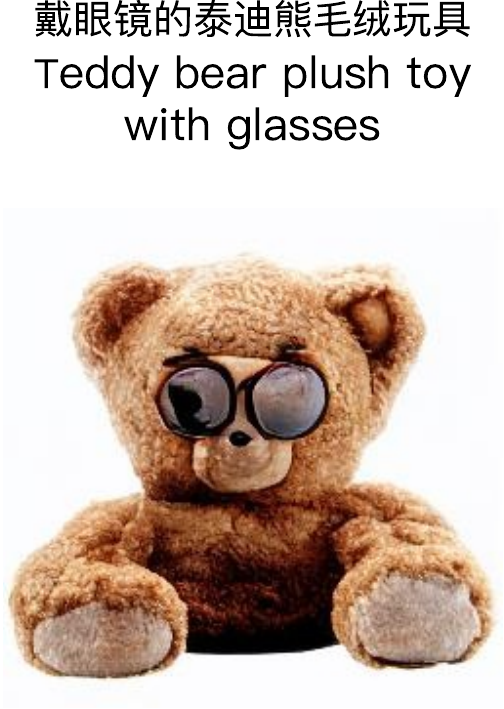}
        \label{fig:t2i_case_openDomain_clx1}
    }
    \subfigure[]{
        \includegraphics[width=0.15\linewidth]{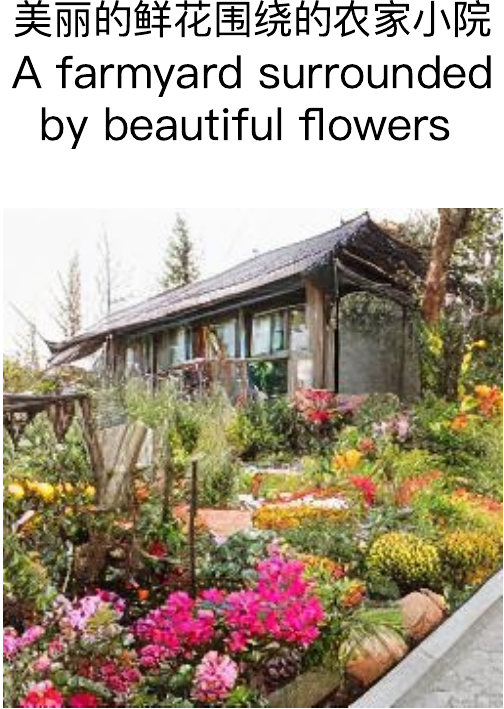}
        \label{fig:t2i_case_openDomain_clx2}
    }
    \subfigure[]{
        \includegraphics[width=0.15\linewidth]{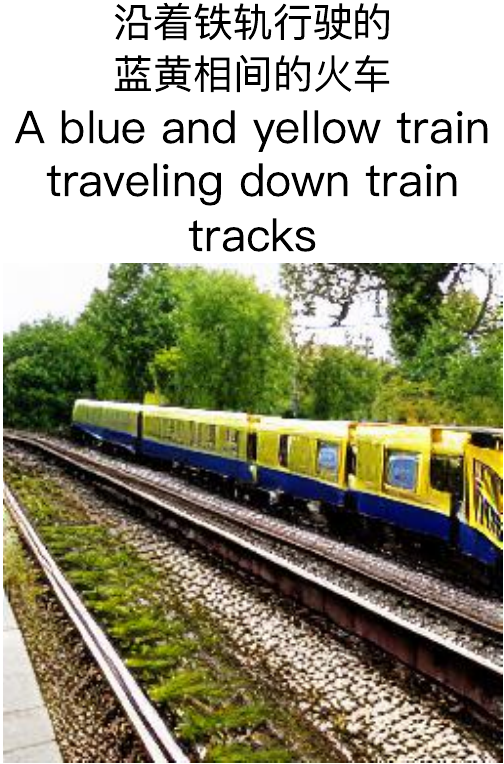}
        \label{fig:t2i_case_openDomain_clx3}
    }
    \subfigure[]{
        \includegraphics[width=0.15\linewidth]{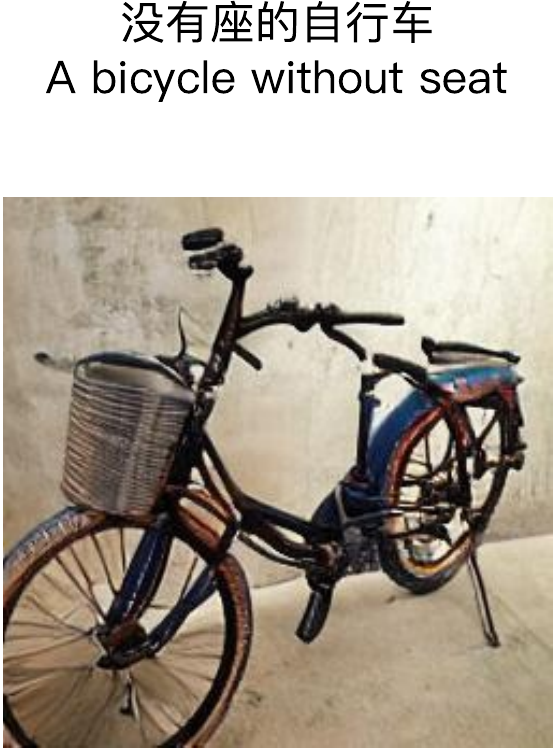}
        \label{fig:t2i_case_openDomain_counterFact1}
    }
    \caption{Example images ERNIE-ViLG generated in zero-shot setting with texts from open domain. Figure \ref{fig:t2i_case_openDomain_sim1}-Figure \ref{fig:t2i_case_openDomain_sim2} show the generated images of simple objects. Figure \ref{fig:t2i_case_openDomain_clx1}-Figure \ref{fig:t2i_case_openDomain_clx3} exhibit generated images of complex scenes with multiple objects. The example of creating image of non-existing objects is displayed in Figure \ref{fig:t2i_case_openDomain_counterFact1}.}
    \label{fig:t2i_case_openDomain}
\end{figure}

\begin{figure}[htb]
    \centering
    \includegraphics[width=0.9\textwidth]{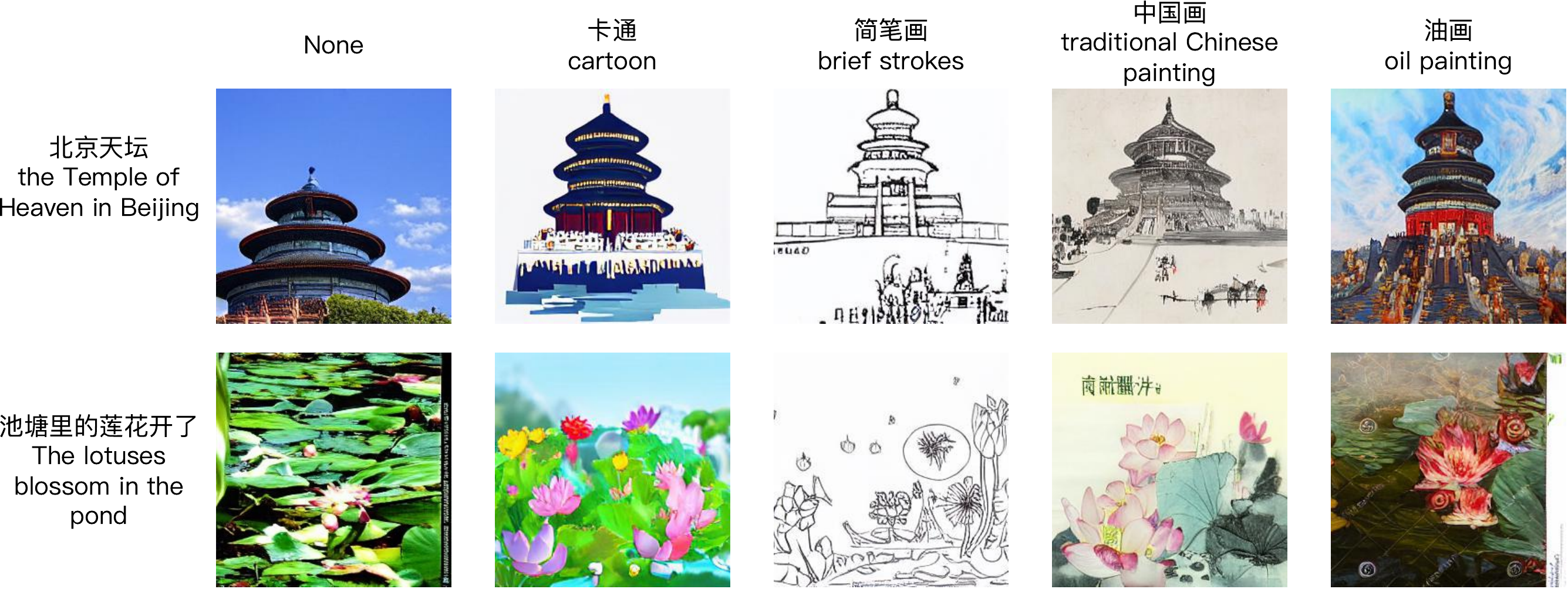}
    \caption{Images of different styles generated by ERNIE-ViLG. "None" indicates not adding any prompts about image style.}
    \label{fig:t2i_case_style}
\end{figure}
\begin{figure}[htb]
    \centering
    \includegraphics[width=0.9\textwidth]{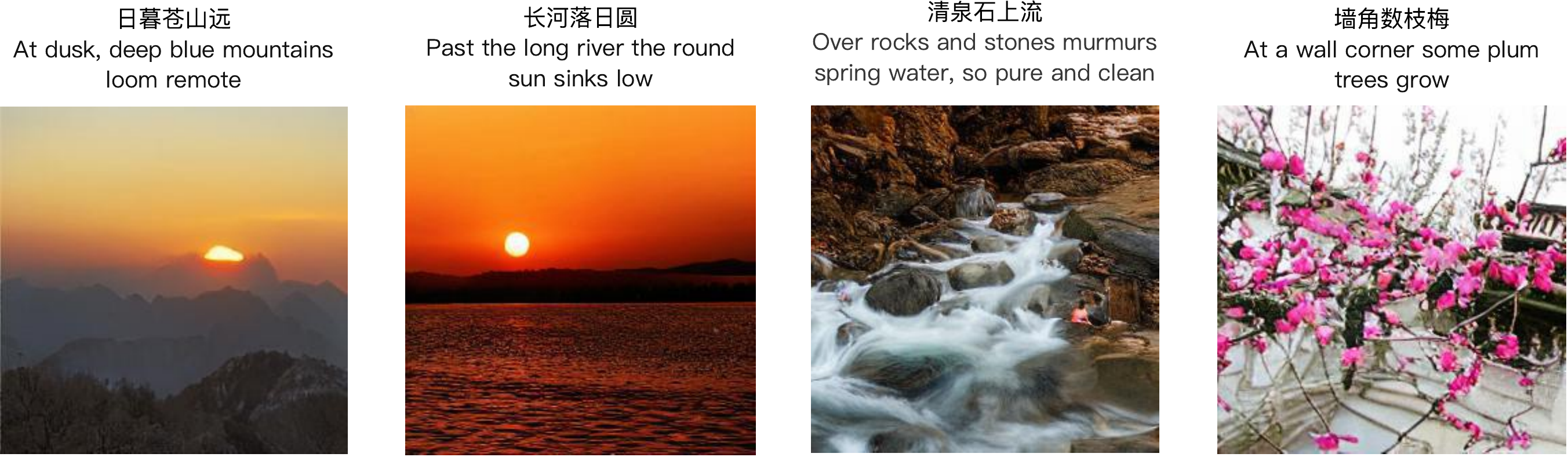}
    \caption{Generated images given Chinese ancient poetry.}
    \label{fig:t2i_case_poetry}
\end{figure}

\paragraph{Qualitative Results}
ERNIE-ViLG has acquired the generation capability for various scenes, from basic objects to complex combinations of objects. 
Some examples are shown in Figure \ref{fig:t2i_case_openDomain}. As can be seen in the examples, ERNIE-ViLG can not only draw entities mentioned in the given text description, but also combine them together with the background in a reasonable way. Surprisingly, we also find two special skills ERNIE-ViLG develops. Firstly, ERNIE-ViLG can generate images of different styles by simply adding text prompts without fine-tuning like CogView does (Figure \ref{fig:t2i_case_style}). Secondly, our model can generate realistic images given Chinese ancient poetry which shows 
promising understanding of brief and abstractive descriptions. Real concepts in the poetry are well-organized and artistic conception is well-described (Figure \ref{fig:t2i_case_poetry}).

\subsection{Image Captioning}
To fully assess the capability of ERNIE-ViLG for image captioning, we carry out automatic evaluations for a fair comparison with previous methods and also evaluate the quality of generated captions through human judgement. 

\paragraph{Automatic Evaluation}
We conduct experiments on two commonly used Chinese image captioning datasets, AIC-ICC \cite{wu2017aicicc} and COCO-CN \cite{li2019cococn}. For AIC-ICC, we fine-tune ERNIE-ViLG on the training split and evaluate on the validation split, following the same practice as \cite{huo2021wenlan}. And for the experiments on COCO-CN, we use the standard train/val/test split and report results on the test split. We adopt the widely used metrics, BLEU@4, METEOR, ROUGE-L and CIDERr, for all the evaluations.

As illustrated in Table \ref{exp-ft-i2t-auto-cococn-result} and Table \ref{exp-ft-i2t-auto-aicicc-result}, ERNIE-ViLG achieves the best results on both datasets. Specifically, compared to the pre-training-based method \cite{huo2021wenlan}, we obtain a significant improvement of 2.1 BLEU@4 on AIC-ICC.

\begin{table}[htb]
  \caption{Evaluation results for image captioning on COCO-CN test split. The results are all calculated at character-level where the results of 'seq-learn' are re-evaluated with their provided predictions.}
  \begin{tabular}{l|cccc}
    \toprule
    Model    & BLEU@4 & METEOR & ROUGE-L & CIDERr \\
    \midrule
    seq-learn \cite{li2019cococn} & 48.4 & 29.5 & 59.2 & 128.4 \\
    \midrule
    ERNIE-ViLG & 50.0 & 31.6 & 60.3 & 138.2 \\
    \bottomrule
  \end{tabular}
  \label{exp-ft-i2t-auto-cococn-result}
  \centering
\end{table}

\begin{table}[htb]
  \caption{Evaluation results for image captioning on AIC-ICC val split.}
  \begin{tabular}{l|cccc}
    \toprule
    Model    & BLEU@4 & METEOR & ROUGE-L & CIDERr \\
    \midrule
    BriVL \cite{huo2021wenlan} & 66.1 & 41.1 & 71.9 & 220.7 \\
    \midrule
    ERNIE-ViLG & 68.2 & 41.7 & 72.5 & 231.4 \\
    \bottomrule
  \end{tabular}
  \label{exp-ft-i2t-auto-aicicc-result}
  \centering
\end{table}

\paragraph{Human Evaluation}
We also carry out a human evaluation campaign to assess the quality of captions generated by ERNIE-ViLG. Given the image and generated predictions, the evaluator is asked to evaluate the caption quality from three perspectives: fluency (whether the caption is fluent or not), relevance (whether the caption is related to the given image or not) and richness (whether the caption adequately describe the whole content of the image or not) with a score from 0 to 2 (higher is better).
We randomly select 200 images from COCO-CN test set and make the predictions from the zero-shot ERNIE-ViLG model and the fine-tuned model. 

\begin{figure}[htb]
    \centering
    \includegraphics[width=0.9\textwidth]{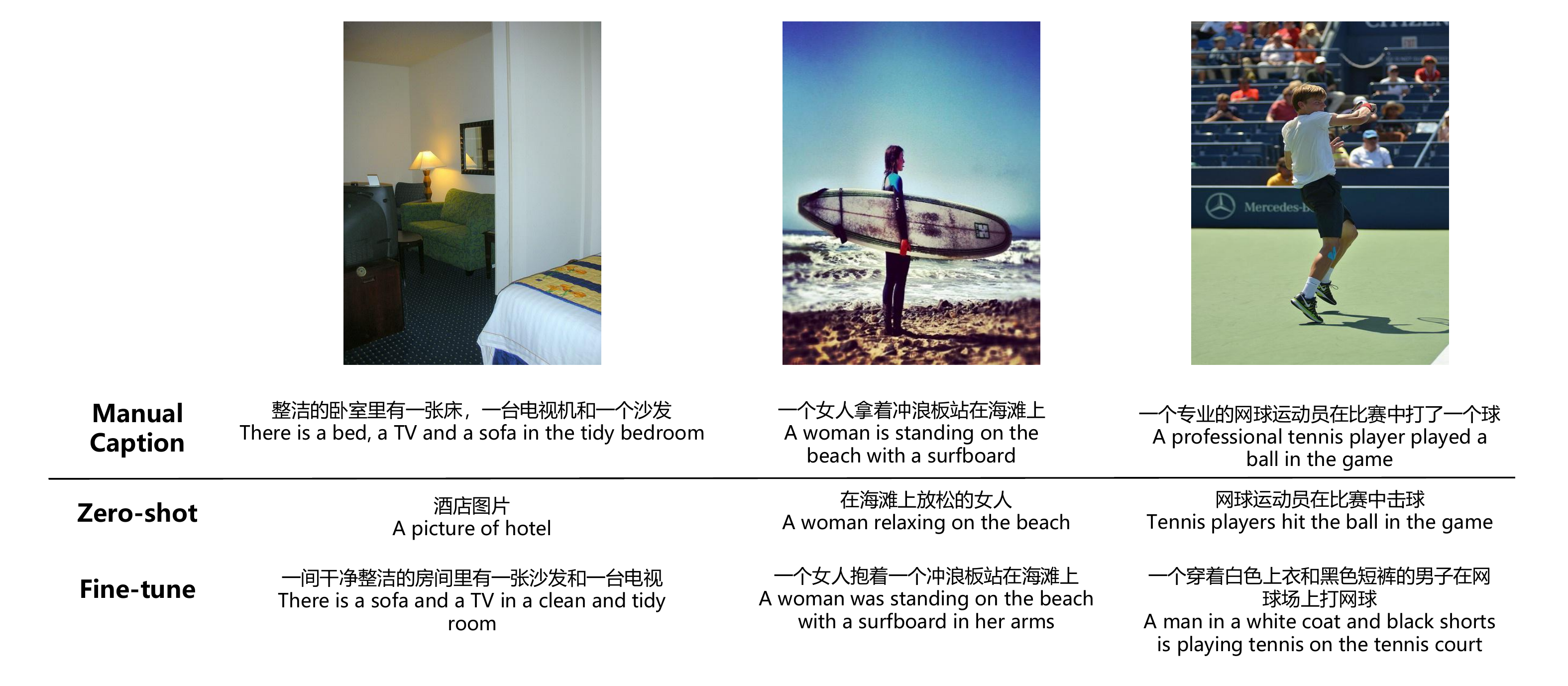}
    \caption{Generated captions on COCO-CN. Although the zero-shot generated captions are various, it is sufficient to describe the image from the semantic perspective.}
    \label{fig:i2t_cases}
\end{figure}

The evaluation results are shown in Table \ref{exp-zero-i2t-human-cococn-result}. The fine-tuned ERNIE-ViLG model obtains an average score of 1.62. We also find that zero-shot ERNIE-ViLG model obtains a high fluency score, close to that of fine-tuned model, but the gap of the relevance score and the richness score between them are significant. We consider that it is due to that the web-crawled pre-training dataset tends to be noisy and most captions are less descriptive while the captions in human-labeled captioning datasets (e.g., COCO-CN) are often descriptive which capture all the details of the image. Some examples are illustrated in Figure \ref{fig:i2t_cases}.

\begin{table}[htb]
  \caption{Human evaluation of the generated captions on the sampling images from COCO-CN test split. }
  \begin{tabular}{l|ccc|c}
    \toprule
    Model    & fluency & relevance & richness & average score\\
    \midrule
    ERNIE-ViLG (zero-shot) & 1.82 & 1.09 & 0.76 & 1.22 \\
    ERNIE-ViLG & 1.96 & 1.47 & 1.43 & 1.62\\
    \bottomrule
  \end{tabular}
  \label{exp-zero-i2t-human-cococn-result}
  \centering
\end{table}

\subsection{Generative Visual Question Answering}
In the literature, the Visual Question Answering task is often converted to a multi-label classification problem over a predefined candidate answer set. However, it is hard to define a closed set of candidate answers in real-world application, which makes this simplification remains a gap between research and practical application. In this paper, we study the open-ended generative Visual Question Answering task, where the model is required to directly generate the answer, given the image and question. An ideal generative model for this task needs to build semantic connections between the image and the question based on visual and linguistic understanding, and generate fluent textual answer. We conduct experiments on a public FMIQA \cite{gao2015fmiqa} dataset, which has freestyle and diversified question-answer annotations. For the evaluation, we conduct both Turing Test \cite{gao2015fmiqa} (a human evaluator is asked to judge whether the answer is generated by a machine or not) and answer quality evaluation (score from 0 to 2, higher is better) by human's judgement. 

We randomly select 200 samples from FMIQA validation split for the evaluations, as there is no official release of the test set. We fine-tune ERNIE-ViLG on the train split and use the fine-tuned model to make predictions for the samples. We also make the evaluation for the human-labeled answers to eliminate the bias of different evaluators compared to \cite{gao2015fmiqa}. The evaluation results can be seen in Table \ref{exp-ft-i2t-human-fmiqa-result}. ERNIE-ViLG achieves a Turing Text passing rate of 78.5\%, which significantly surpasses that of mQA \cite{gao2015fmiqa} and an answer score of 1.495 which clearly verify that our model has captured the semantic alignments across vision and language modalities. 

\begin{table}[htb]
  \caption{Human evaluation on FMIQA. * means the evaluations on manual captions for our samples.}
  \begin{tabular}{l|c|c}
    \toprule
    \multirow{2}{*}{Model}    & Turing Test & Quality Evaluation\\
                       & passing rate(\%) & avg.score\\
    \midrule
    Human annotations from \cite{gao2015fmiqa}    & 94.8          & 1.918 \\
    mQA \cite{gao2015fmiqa}      & 64.7          & 1.454 \\
    \midrule
    Human annotations *              & 97.5           & 1.870 \\
    ERNIE-ViLG              & 78.5           & 1.495 \\
    \bottomrule
  \end{tabular}
  \label{exp-ft-i2t-human-fmiqa-result}
  \centering
\end{table}

\section{Analysis} \label{section:end2endexp}
To evaluate the benefits our proposed end-to-end text-to-image synthesis method brings, we conduct experiments with a lite version of transformer as the image discrete representation generator, which has about 300 million parameters (24 transformer layers with 1024 hidden units and 16 attention heads).
We utilize dVAE \cite{ramesh2021zero}, rather than VQGAN, to discretize the image to visual sequence and build the image from the visual tokens, considering the instability of the training process of GAN. For the reconstructor, we use the same network architecture as dVAE. We develop a multi-task learning process which assigns equal weights to the generation loss and the reconstruction loss. The models are trained on a merged dataset of CC and CC12M and the evaluation is carried out on the MS-COCO validation set in  zero-shot setting. 

We compare the results of our end-to-end training method with the two-stage pipeline baseline as shown in Table \ref{tab:end2end-gen-result}. For the two-stage pipeline, we train a text-to-image generator and use the decoder of dVAE directly as the reconstructor. The Two-stage G(R) refers to the separately trained generator(reconstructor), and the end-to-end G(R) refers to the end-to-end trained generator(reconstructor). Our end-to-end method achieves a significant FID improvement of 1.5 compared to the two-stage pipeline. 
We find that combining the end-to-end trained generator (end-to-end G) and dVAE decoder (two-stage R) also brings a FID improvement of 0.9 compared to that of two-stage pipeline, but falls behind compared to the end-to-end methods. This indicates our proposed end-to-end method can improve both the performance of the generator (two-stage G \& two-stage R vs end-to-end G \& two-stage R) and the reconstructor (end-to-end G \& two-stage R vs end-to-end G \& end-to-end R).  

We also input visual sequence of real images discretized by dVAE(gold image sequence) to the two reconstructors for comparison. Experimental results (the last two lines in Table \ref{tab:end2end-gen-result}) show that the end-to-end trained reconstructor has more obvious advantage in the reconstruction from real image discrete representation. We consider that end-to-end training will be more effective on ERNIE-ViLG with 10 billion parameters, for the image discrete representation generated by more capable generator is much closer to the real distribution, and hidden embeddings of larger model provides more useful features for the reconstructor. Due to the instability of the training of both GAN and large-scale generative model, we haven't used end-to-end training for our 10-billion parameter model based on VQGAN. We will address the instability issue for future work and improve the 10-billion parameter ERNIE-ViLG through end-to-end training.

\begin{table}[htb]
  \caption{Comparison of the two-stage pipeline and end-to-end training.}
  \begin{tabular}{c|c|c}
    \toprule
    Generator & Reconstructor  & FID $\downarrow$ \\
    \midrule
    Two-stage G & Two-stage R & 41.4 \\
    End-to-end G & Two-stage R &  40.5 \\
    End-to-end G & End-to-end R & 39.9 \\
    Gold image sequence & Two-stage R & 21.7 \\
    Gold image sequence & End-to-end R & 18.6 \\
    \bottomrule
  \end{tabular}
  \label{tab:end2end-gen-result}
  \centering
\end{table}

\section{Conclusion}
We propose ERNIE-ViLG to unify the bidirectional image-text generation tasks in one single generative model and present an end-to-end training method for the text-to-image synthesis. Pre-trained on a large-scale image-text dataset, ERNIE-ViLG captures the capabilities of bidirectional Vision-Language Generation and achieves superior performance on various cross-modal generation tasks including text-to-image synthesis, image captioning and generative Visual Question Answering. Overall, our model advances the unified pre-training for both the image-to-text and text-to-image generation tasks further. 

\bibliographystyle{unsrt}
\bibliography{references}  

\appendix
\section{Dataset for human evaluation of text-to-image synthesis}
\label{sec: t2i-human-eval-dataset}
\begin{table}[htb]
 \caption{Statistics of the dataset for human evaluation.}
  \centering
  \begin{tabular}{l|l|c}
    \toprule
         & Description Angle & Number \\
    \midrule
    1   &  MS-COCO & 102 \\  \midrule
    2   &  anthropomorphic animal / cartoon characters & 49 \\  \midrule
    3   &  geography & 50 \\    \midrule
    4   &  multi-object + attribute description + relationship description & 68 \\  \midrule
    5   &  single-object + attribute description & 56 \\    \midrule
    6   &  counter fact   & 54 \\ \midrule
    7   &  different view angles    & 43 \\ \midrule
    8   &  different styles & 43 \\ \midrule
    9   &  different time, different scenes & 35 \\ \midrule
  \end{tabular}
  \label{tab: app-zero-t2i-human-data}
\end{table}
The dataset for human evaluation of text-to-image synthesis has 500 texts that are collected from 9 different aspects. Details of this dataset is shown in Table \ref{tab: app-zero-t2i-human-data}. 102 texts are randomly selected from MS-COCO's validation set, while the others are manually designed. 

\end{document}